\documentclass{article}

\usepackage{arxiv}

\usepackage[utf8]{inputenc} 
\usepackage[T1]{fontenc}    
\usepackage{hyperref}       
\usepackage{url}            
\usepackage{booktabs}       
\usepackage{amsfonts}       
\usepackage{nicefrac}       
\usepackage{microtype}      
\usepackage{graphicx}
\usepackage{natbib}
\usepackage{doi}

\title{Exploring the Word Sense Disambiguation Capabilities of Large Language Models}

\author{ \href{https://orcid.org/0000-0002-0545-1105}{\includegraphics[scale=0.06]{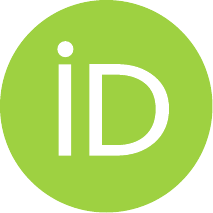}\hspace{1mm}Pierpaolo Basile}\\
	Department of Computer Science\\
	University of Bari Aldo Moro\\
	Bari, Italy \\
	\texttt{pierpaolo.basile@uniba.it} \\
    \And
    \href{https://orcid.org/0000-0002-1438-280X}{\includegraphics[scale=0.06]{orcid.pdf}\hspace{1mm}Lucia Siciliani}\\
	Department of Computer Science\\
	University of Bari Aldo Moro\\
	Bari, Italy \\
	\texttt{lucia.siciliani@uniba.it}
    \AND
    \href{https://orcid.org/0009-0006-9670-9998}{\includegraphics[scale=0.06]{orcid.pdf}\hspace{1mm}Elio Musacchio}\\
	Department of Computer Science\\
	University of Pisa\\
	Pisa, Italy \\
	\texttt{elio.musacchio@phd.unipi.it}
     \And
     \href{https://orcid.org/0000-0001-6883-1853}{\includegraphics[scale=0.06]{orcid.pdf}\hspace{1mm}Giovanni Semeraro}\\
	Department of Computer Science\\
	University of Bari Aldo Moro\\
	Bari, Italy \\
	\texttt{giovanni.semeraro@uniba.it}
}

\hypersetup{
pdftitle={Exploring the Word Sense Disambiguation Capabilities of Large Language Models},
pdfsubject={Computer Science, Computation and Language},
pdfauthor={Pierpaolo Basile, Lucia Siciliani, Elio Musacchio, Giovanni Semeraro},
pdfkeywords={Large Language Models, Word Sense Disambiguation, Lexical Semantics, Artificial Intelligence},
}

\begin{document}

\maketitle

\begin{abstract}
Word Sense Disambiguation (WSD) is a historical task in computational linguistics that has received much attention over the years. However, with the advent of Large Language Models (LLMs), interest in this task (in its classical definition) has decreased.
In this study, we evaluate the performance of various LLMs on the WSD task. We extend a previous benchmark (XL-WSD) to re-design two subtasks suitable for LLM: 1) given a word in a sentence, the LLM must generate the correct definition; 2) given a word in a sentence and a set of predefined meanings, the LLM must select the correct one. The extended benchmark is built using the XL-WSD and BabelNet. The results indicate that LLMs perform well in zero-shot learning but cannot surpass current state-of-the-art methods. However, a fine-tuned model with a medium number of parameters outperforms all other models, including the state-of-the-art.
\end{abstract}

\section{Introduction}

The Word Sense Disambiguation (WSD) task \cite{ide1998introduction,navigli2009word,bevilacqua2021recent} has a long tradition in computational linguistics.
The most used definition of WSD is selecting the correct meaning for a word occurrence in a text from a set of possible meanings provided by a sense inventory.
The classical definition requires the existence of a sense inventory that provides for each word a list of possible meanings. This inventory can be a dictionary, a thesaurus, or a semantic network such as WordNet \cite{miller1995wordnet}.
However, since the existence of a sense inventory can be a limit in particular contexts, the task of Word Sense Discrimination/Induction was introduced.
In this case, the task is to infer the different word usages by clustering the word occurrences according to their meaning; in this way, occurrences that share the same meaning are grouped in the same cluster, and the final set of clusters corresponds to the differing meanings of that word.  

In the past, several techniques were exploited to solve both tasks. In particular, the WSD methodologies evolved according to advances in Artificial Intelligence and Machine Learning.
The first period was characterized by using rules-based systems followed by knowledge-based approaches when digital sense inventories became available.
With the availability of digital corpora, supervised approaches were introduced to take advantage of manually annotated data.
With the advent of the web, large corpora and large knowledge graphs automatically extracted for the web revolutionized supervised and knowledge-based approaches.
A new set of approaches was proposed when language models based on the transformers' architecture \cite{vaswani2017attention} were introduced.
The ability of these models to represent words in context through dense vectors opens new possibilities for the disambiguation and discrimination of word meanings.

The more recent novelty that revolutionized computational linguistics is the introduction of Large Language Models (LLMs).
Essentially, an LLM is based on the transformer architecture trained on vast amounts of text data to understand and generate human-like language.
LLMs have proven their ability to solve different tasks in a zero-shot or few-shot setting without using specific training data.
However, it is also possible to fine-tune an LLM on specific tasks using training data.
The capability of LLMs to solve several tasks without training suggests an intrinsic ability to understand the semantics behind the language.
The impressive results achieved by LLMs might make us lose sight of or underestimate the problem of automatic disambiguation of meaning.
In this work, we want to measure how state-of-the-art LLMs can solve the WSD task to understand if the model somehow stores knowledge about word meanings.
The WSD task must be redesigned considering the generative abilities of LLMs. Therefore, we extend a previous benchmark with two subtasks suitable for testing an LLM.
We re-design the WSD task in two ways: 1) the model is tested in generating the definition of a word in a sentence; 2) the model is evaluated in selecting the correct meaning of a word in a sentence from a predefined set of possible choices following a multiple choices paradigm often used to evaluate LLMs.

Our study considers only \textbf{open} LLMs with a different number of parameters and several languages: English, Spanish, French, Italian and German.
The performance is evaluated according to a gold standard, considering the quality of the generated definition (sub-task 1) and the correctness of the selected sense (sub-task 2).

The main contributions of our work are 1) the extension of an existing multilingual benchmark for testing and training LLMs in the context of the WSD task; 2) an extensive evaluation of open state-of-the-art LLMs; 3) the release of several fine-tuned models trained on our dataset.

The paper is structured as follows: Section \ref{sec:related} discusses related works that leverage LLMs for solving WSD; Section \ref{sec:dataset} provides details about the benchmark used to evaluate LLMs in the WSD task, while Section \ref{sec:methodology} describes the methodology used to generate and select answers exploiting LLMs.
Results are discussed in Section \ref{sec:evaluation}, and final remarks are reported in Section \ref{sec:conclusions}.

\section{Related Work}\label{sec:related}

Transformer-based language models are widely used for solving the WSD task.
A deep overview is proposed in \cite{loureiro-etal-2021-analysis}.
BERT \cite{devlin2018bert} and its variations excel in understanding context-sensitive semantic nuances, making them dominant in evaluation benchmarks.
The authors find that BERT-like models can accurately distinguish between different word senses, even with limited examples for each sense.
The analysis also shows that while language models can nearly solve coarse-grained noun disambiguation under ideal conditions (ample training data and resources), such scenarios are rare in real-world applications, leaving significant challenges.
Moreover, the article compares two WSD strategies: fine-tuning and feature extraction.
The authors conclude that feature extraction is more robust, especially in dealing with sense bias and when training data is limited.
Notably, averaging contextualized embeddings as a feature extraction method is effective, performing well even with just few training sentences per word sense. 
However, no works about the usage of recent LLMs for WSD are proposed.
In our work, we try to investigate the ability of LLMs in a zero-shot setting without any training data and considering a fine-grained sense inventory.
Moreover, we also propose an analysis of a fine-tuned LLM when training data are available.

Another interesting analysis of WSD approaches based on BERT-like models is proposed in \cite{bevilacqua2021recent}.
This work analyzes several WSD approaches including ones that leverage language-models both for extracting contextual-embeddings used as features and as starting point for training a supervised model on sense-annotated data.
This paper is strongly related to our work because it provides an extensive evaluation of the same dataset we considered.

Another interesting work is the one proposed by \cite{cabiddu2023comparing}, where several language models, including large models such as GPT and GPT-2, are evaluated in three behavioural experiments used to measure children's sense disambiguation capabilities.
The study is interesting because it tries to compare how semantics is perceived by children and how it is represented in transformer-based models.
The authors find a model bias with respect to the most dominant meaning and a negative correlation between the training size and the model performance. However, the authors limit their analysis to this dataset, which is very specific.

Recently, some works have investigated the usage of LLMs in solving the WSD task.
For example, \cite{kibria2024functional} assessed the WSD accuracy of LLMs across eight datasets using a multiple-choice question task. More recently, \cite{yae2025leveraging} extended this analysis by evaluating LLM performance on single-choice questions and investigating the influence of varying parameter sizes on disambiguation accuracy.
In \cite{kritharoula-etal-2023-large}, the authors combine transformer-based methods for multimodal retrieval and LLMs to solve the task of Visual WSD (VWSD).
VWSD is a novel task that aims to retrieve an image among a set of candidates that better represents the meaning of an ambiguous word within a given context. LLMs are used as knowledge bases to enhance the textual context and resolve ambiguity related to the target word. ARPA, the system proposed in \cite{papastavrou2024arpa}, achieves SOTA performance in VWSD by leveraging LLMs to generate word embeddings.

In conclusion, several works exploited transformer-based architectures similar to BERT for WSD essentially in two ways: 1) extraction of contextual embeddings used as features; 2) supervised models that fine-tune the language model on sense-annotated data. However, few works have exploited recent decoder-only LLMs as word sense disambiguators in a zero-shot setting (completely unsupervised) or as a base for further fine-tuning on annotated data. Our work tries to fill this gap by considering the more recent and state-of-the-art open LLMs. 

\section{The Benchmark}\label{sec:dataset}

To evaluate LLMs on the WSD task, we need a sense-annotated corpus, i.e., a collection of sentences in which each word is tagged with its correct meaning taken from a sense inventory.
For this reason, we also require a sense inventory that provides the set of possible meanings for each word.
Therefore, our benchmark requires both a multilingual corpus and a multilingual sense inventory.

We will introduce some formal notations before delving into the description of the benchmark construction.
Given a sentence $S_k$ and one of its word occurrences $w_i$, we define $L_i$ as the list of possible meanings of $w_i$ and $m_j \in L_i$, the meaning assigned to $w_i$. Each meaning has several glosses, one for each language taken into account, and we use $m_{j, lang} \in L_i$ to refer to it.
In our case, $lang \in \{en, it, es, fr, de\}$.
Starting from the multilingual sense-annotated corpus and the corresponding sense inventory, we need a strategy for building two types of prompts for testing LLMs.

The first prompt aims to assess the ability of the LLM to generate an accurate definition of a word within a specific sentence.
For each sense annotated word occurrence, we create the prompt \cite{liu2023pre} in Table \ref{tab:promptGen} for each language. The table reports only the prompt for English; the others are provided in the Appendix \ref{sec:appendixB}.
We also store the correct definition $m_j$ in the benchmark in a field called \texttt{output}.

\begin{table}[!h]
\centering
\begin{tabular}{|p{7cm}|}
\hline
\multicolumn{1}{|c|}{\textbf{Prompt template (generation)}} \\
\hline
Give a brief definition of the word "$w_i$" in the sentence given as input. Generate only the definition. Input: "$S_k$"  \\
\hline
\multicolumn{1}{|c|}{\textbf{English prompt}}  \\
\hline
Give a brief definition of the word "art" in the sentence given as input. Generate only the definition. Input: "The art of change-ringing is peculiar to the English, and, like most English peculiarities, unintelligible to the rest of the world." \\
\hline
\end{tabular}\caption{Prompt for the generation benchmark.}\label{tab:promptGen}
\end{table}

While constructing the prompt, we need to manage the cases in which a word $w_i$ occurs more than once in the sentence $S_k$. In these cases, we change the prompt as follows: ``\textit{Give a brief definition of the $x$ occurrence of the word "$w_i$"}...'', where $X = \{first, second, third, fourth, fifth\}$ and $x \in X$. We exclude cases where the word occurs more than six times, and we translate the set $X$ according to each language.

The goal of the second kind of prompt is to evaluate the LLM's ability to select the correct sense from a set of predefined possibilities following a multiple-choice paradigm. In this case, we exploit the list of all possible meanings $L_i$. In particular, from $L_i$, we remove all the annotated meanings\footnote{In the sense-annotated corpus, a word occurrence can be annotated with more than one correct meaning.} and obtain the set $C_i$.
Then, we randomly add to $C_i$ one of the correct meanings; in this way, $C_i$ contains only one correct sense.
For each occurrence of a sense-annotated word in the corpus, we create the prompt in Table \ref{tab:promptSel} for each language. The table reports only the prompt for English; the others are provided in the Appendix \ref{sec:appendixB}.
Additionally, we store the identifier (i.e. the option's number) corresponding to the correct answer in a field called \texttt{output}.

\begin{table}[!h]
\centering
\begin{tabular}{|p{7cm}|}
\hline
\multicolumn{1}{|c|}{\textbf{Prompt template (multiple choice)}} \\
\hline
Given the word "$w_i$" in the input sentence, choose the correct meaning from the following:
$C_i$.
Generate only the number of the selected option.  \\
\hline
\multicolumn{1}{|c|}{\textbf{English prompt}}  \\
\hline
Given the word "art" in the input sentence, choose the correct meaning from the following:\\
1) Photographs or other visual representations in a printed publication\\
2) A superior skill that you can learn by study and practice and observation\\
3) The products of human creativity; works of art collectively\\
4) The creation of beautiful or significant things.\\
Generate only the number of the selected option.\\
Input: "The art of change-ringing is peculiar to the English, and, like most English peculiarities, unintelligible to the rest of the world." \\
\\\hline
\end{tabular}\caption{Prompt for the multiple choice benchmark.}\label{tab:promptSel}
\end{table}

We also manage the case where the word $w_i$ occurs more than once by modifying the prompt as in the first benchmark.
Moreover, given that the model is asked to choose among different options in this benchmark, we need to manage cases in which the size of $C_i$ is less than two.
In these cases, we remove the occurrence from the dataset. Monosemic words are not considered in the construction of both tasks\footnote{For the first benchmark based on definition generation, it is also possible to consider monosemic words. We exclude this hypothesis since we want to test LLMs in the case of polysemy.}.

We use XL-WSD \cite{pasini-etal-xl-wsd-2021} as our sense-annotated corpus. This dataset serves as a cross-lingual evaluation benchmark for the WSD task, featuring sense-annotated development and test sets in 18 languages from six different linguistic families. Additionally, it includes language-specific training data, making it highly useful for evaluating WSD performance in a multilingual context.
As stated previously, this study is focused on five languages: English, Italian, Spanish, French, and German.
The sense inventory adopted in XL-WSD is BabelNet \cite{navigli-ponzetto-2010-babelnet}.
However, not all senses in BabelNet have a gloss for each of the chosen languages.
For this reason, we build two versions of the dataset: \textbf{without translation} in which we consider only the word occurrences that have glosses in BabelNet for each language, and \textbf{with translation} in which English glosses\footnote{The English gloss is always available.} are automatically translated when they are not available in BabelNet for a particular language. We use the 1.3B variant of the Meta NLLB-200 model\footnote{\url{https://huggingface.co/facebook/nllb-200-1.3B}} for the translation. We selected this translation model because it has a good performance and computational cost trade-off. Moreover, it is open.

The Table \ref{tab:datasetInfo} reports the number of instances for each kind of task. Statistics for each language are reported in Appendix \ref{sec:appendixA}.

\begin{table}[!htb]
\centering
\begin{tabular}{lll}
\hline
\multicolumn{3}{c}{\textbf{Without translation}}                                                            \\ \hline
\multicolumn{1}{l|}{}             & \multicolumn{1}{c}{Generation} & \multicolumn{1}{c}{Multiple-choice} \\ \cline{2-3} 
\multicolumn{1}{l|}{Training set} & 1,204,430                              & 861,791                             \\
\multicolumn{1}{l|}{Test set}     & 10,480                              & 9,847                             \\ \hline
\multicolumn{3}{c}{\textbf{With translation}}                                                               \\ \hline
\multicolumn{1}{l|}{}             & \multicolumn{1}{c}{Generation} & \multicolumn{1}{c}{Multiple-choice} \\ \cline{2-3} 
\multicolumn{1}{l|}{Training set} & 1,451,650                              & 1,170,921                             \\
\multicolumn{1}{l|}{Test set}     & 11,473                             & 11,168                            
\end{tabular}\caption{Task statistics: number of instances.}\label{tab:datasetInfo}
\end{table}

\section{Methodology}\label{sec:methodology}

We follow two distinct methodologies to evaluate LLMs in solving the WSD task.
In the first approach, known as (\textbf{zero-shot}), we directly prompt a selection of open LLMs with a varying number of parameters, without any task-specific training, to assess their inherent ability to solve the WSD problem.
In the second approach, we \textbf{fine-tune} an open LLM with a small number of parameters.
Our aim is twofold: 1) testing existing open models in solving the disambiguation task without additional training and 2) determining whether a model with a small number of parameters can solve the disambiguation task with proper fine-tuning.
We select a model with a small number of parameters to allow the training on more accessible hardware.

\subsection{Zero-shot}\label{subsec:zeroshot}

The methodology of this approach is straightforward. 
We select a set of open LLMs with a different number of parameters and directly prompt them using the benchmarks described in Section \ref{sec:dataset}.
Then, we measure the quality of the generated definitions and the accuracy in selecting the correct sense from the predefined alternatives.
We perform two separate evaluations: the former involves only the original glosses, while the latter also contains the machine-translated ones.

For prompting the models, we use a cloud service\footnote{together.ai: \url{https://www.together.ai/}. We spend about 15\$ for performing all the experiments.}.
We consider the following LLMs: Llama-3.1-instruct-8B \cite{dubey2024llama}, Mistral-instruct-7B-v03 \cite{jiang2023mistral7b}, Gemma2-9B \cite{gemmateam2024gemma2improvingopen}, Llama-3.1-instruct-70B, Qwen2-72B-Instruct \cite{yang2024qwen2} and Llama-3.1-instruct-405B.

All models are tested using a greedy search approach as an inference strategy to avoid variability over different runs.

\subsection{Fine-tuning}\label{subsec:finetuning}

We use \textsc{Llama 3.1 8B Instruct} as the base model to fine-tune.
The \textsc{LLaMA 3.1} family of models has been designed and trained with multilinguality in mind, by properly balancing the languages in the training mixture.
Therefore, the \textsc{LLaMA 3.1} models are already skilled in understanding and generating text in multiple languages.
Specifically, they support the following languages: English, German, French, Italian, Portuguese, Hindi, Spanish, and Thai.
All languages considered in this work are also natively supported by \textsc{Llama 3.1 8B Instruct}, making it the ideal starting point.

We use a full-parameter training approach for the fine-tuning strategy, using \textit{DeepSpeed ZeRO 3}\footnote{\url{https://github.com/microsoft/DeepSpeed}} \citep{rajbhandari2020zero} for parallelization.
We rely on a compute node consisting of four A100 64GB VRAM GPUs. We use the \textsc{LLaMA 3.1} instruct template to format the prompt without any system message. We use a maximum sequence length of $512$, discarding examples exceeding this value. The value was selected after studying the number of tokens in the training sets with and without translation (also considering the tokens added by the instruction formatting). In all cases, the 95th percentile of the number of tokens was less than $512$. Other relevant hyperparameters are reported in Table \ref{tab:hyperparameters}.

\begin{table}[ht]
    \large
    \centering
    \begin{tabular}{l|c}
         \textbf{Parameter} & \textbf{Value} \\
         \hline
         batch size & 512 \\ 
         lr & 4e-5 \\
         lr warmup ratio & 0.00 \\
         weight decay & 0 \\
         epochs & 1 \\
         optimizer & AdamW 
    \end{tabular}
    \caption{Hyperparameters for fine-tuning}
    \label{tab:hyperparameters}
\end{table}

\section{Evaluation}\label{sec:evaluation}

This section reports the results of both evaluations: the first involves several LLMs in a zero-shot setting, while the second is based on a fine-tuned model based on \textsc{Llama 3.1 8B Instruct}.

We select two metrics to evaluate the quality of generated definitions: 1) RougeL refers to the overlap of longest co-occurring in sequence n-grams between the reference text and the generated one; 2) BERTscore \cite{zhang2019bertscore} exploits the pre-trained contextual embeddings from BERT and matches words in reference and generated definition by cosine similarity.
RougeL gives us an idea of the syntactic similarity, while the BERTscore measures semantic coherence.
The BERTscore is necessary since the LLM can generate a most extended or lexically different definition that is semantically correct. For the computation of the BERTscore we use an English BERT model for evaluating the English part of the dataset, while we exploit a multilingual version of BERT for the other languages.
We use accuracy to measure the LLM's ability to select the correct sense from a set of alternatives.

We explore different settings since we use machine translation to create the missing glosses for all the languages.
We perform zero-shot experiments on two benchmark subsets: 1) without machine translation and 2) with machine translation.
We perform four runs when a fine-tuned model is involved, considering that translated glosses can also be used during the fine-tuning. Table \ref{tab:ftconf} shows all the combinations and the corresponding labels used to reference each configuration.
We perform only two runs for the zero-shot evaluation since we do not consider training data and need to test on two sub-sets: one with translation and one without translation.

\begin{table}[!htb]
\centering
\begin{tabular}{l|l|c}
\textbf{Training data} & \textbf{Test data} & \textbf{Label} \\ \hline
Without translation    & Without translation           & FF             \\
Without translation    & Machine translated         & FT             \\
Machine translated             & Without translation           & TF             \\
Machine translated             & Machine translated         & TT            
\end{tabular}\caption{Different evaluation settings according to the kind of glosses: original glosses in BabelNet and machine translation of missing glosses.}\label{tab:ftconf}
\end{table}

\subsection{Zero-shot results}

In this section, we report the zero-shot evaluation results. Table \ref{tab:resZero} shows results without machine translation, while Table \ref{tab:resZeroMT} with translation.
Gemma2-9B is the best medium model in the multiple-choice task (accuracy) for all languages without translation. We can observe similar performance between llama3.1-8B and Gemma2-9B in terms of the quality of generated definitions. 
Llama3.1-8B achieves the better RougeL score for all languages, while BERTscores are similar to Gemma2-9B. Mistral-7B achieves the worst results, but it is also the smallest model.

\begin{table*}[!htb]
\small
\centering
\begin{tabular}{l|rrr|rrr|rrr}
   & \multicolumn{3}{c|}{Llama3.1 8B-Instruct}                                                           & \multicolumn{3}{c|}{Mistral 7B-Instruct}                                                            & \multicolumn{3}{c}{Gemma2 9B-Instruct}                                                             \\ \cline{2-10} 
   & \multicolumn{1}{c}{RougeL} & \multicolumn{1}{c}{BERTscore} & \multicolumn{1}{c|}{Accuracy} & \multicolumn{1}{c}{RougeL} & \multicolumn{1}{c}{BERTscore} & \multicolumn{1}{c|}{Accuracy} & \multicolumn{1}{c}{RougeL} & \multicolumn{1}{c}{BERTscore} & \multicolumn{1}{c}{Accuracy} \\ \hline
EN & \textbf{.2260}                     & .8638                        & .5587                        & .1149                     & .8314                        & .6171                        & .2116                     & \textbf{.8650}                        & \textbf{.6762}                       \\
IT & \textbf{.1363}                     & .6985                        & .4604                        & .0747                     & .6532                        & .5324                        & .1221                     & \textbf{.6986}                        & \textbf{.5840}                       \\
ES & \textbf{.1811}                     & \textbf{.7262}                        & .5802                        & .1408                     & .6872                        & .5898                        & .1570                     & .7158                        & \textbf{.6503}                       \\
FR & \textbf{.1901}                     & \textbf{.7247}                        & .5090                        & .1437                     & .6888                        & .6290                        & .1208                     & .6815                        & \textbf{.6493}                       \\
DE & \textbf{.1586}                     & \textbf{.7050}                        & .6217                        & .1101                     & .6808                        & .6130                        & .1091                     & .6791                        & \textbf{.6826}                       \\ \hline
   & \multicolumn{3}{c|}{Llama 3.1 70B-Instruct}                                                         & \multicolumn{3}{c|}{Qwen2-72B-Instruct}                                                    & \multicolumn{3}{c}{Llama 3.1 405B-Instruct}                                                        \\ \cline{2-10} 
   & \multicolumn{1}{c}{RougeL} & \multicolumn{1}{c}{BERTscore} & \multicolumn{1}{c|}{Accuracy} & \multicolumn{1}{c}{RougeL} & \multicolumn{1}{c}{BERTscore} & \multicolumn{1}{c|}{Accuracy} & \multicolumn{1}{c}{RougeL} & \multicolumn{1}{c}{BERTscore} & \multicolumn{1}{c}{Accuracy} \\ \hline
EN & \textbf{.2437}                     & .8654                        & .7520                        & .1670                     & .8455                        & .7370                        & .2393                     & \textbf{.8669}                        & \textbf{.7532}                       \\
IT & .1439                     & .7018                        & .6298                        & .1131                     & .6773                        & \textbf{.6396}                        & \textbf{.1524}                     & \textbf{.7072}                        & .6259                       \\
ES & .1900                     & .7231                        & .7012                        & .1749                     & .7096                        & \textbf{.7214}                        & \textbf{.1915}                     & \textbf{.7297}                        & \textbf{.7214}                       \\
FR & .1713                     & .7054                        & .7059                        & .1596                     & \textbf{.7057}                        & \textbf{.7624}                        & \textbf{.1751}                     & .7003                        & .7149                       \\
DE & \textbf{.1454}                     & \textbf{.6991}                        & .7739                        & .1276                     & .6955                        & .7652                        & .1343                     & .6894                        & \textbf{.7957}                      
\end{tabular}\caption{Zero-shot results without machine translation.}\label{tab:resZero}
\end{table*}

As expected, larger LLMs provide better results than the medium counterparts.
Interestingly, Llama3.1-70B and Llama3.1-405B provide similar results despite the significant difference in the number of parameters.
While Llama3.1-405B achieves the best accuracy for English, its performance is only slightly better than that of Llama3.1-70B. Notably, Qwen2-72B-Instruct achieves good results for Spanish and French.

Results with machine translation are shown in Table \ref{tab:resZeroMT}.
The results are pretty similar to those of the experiments without machine translation.
Among medium-sized models, Gemma2-9B provides better accuracy, but Llama3.1-8B provides the best quality in the generation.
Llama3.1-405B achieves the best accuracy and generation quality for larger models, although Llama3.1-70B occasionally provides similar or better performance. Only for Italian, the Qwen2-72B model has the best accuracy.
These findings indicate that machine translation does not significantly affect the performance behaviour between LLMs. However, we observe a general decrease in performance across all LLMs and metrics.

\begin{table*}[!htb]
\small
\centering
\begin{tabular}{l|rrr|rrr|rrr}
   & \multicolumn{3}{c|}{Llama3.1 8B-Instruct}                                                           & \multicolumn{3}{c|}{Mistral 7B-Instruct}                                                            & \multicolumn{3}{c}{Gemma2 9B-Instruct}                                                             \\ \cline{2-10} 
   & \multicolumn{1}{c}{RougeL} & \multicolumn{1}{c}{BERTscore} & \multicolumn{1}{c|}{Accuracy} & \multicolumn{1}{c}{RougeL} & \multicolumn{1}{c}{BERTscore} & \multicolumn{1}{c|}{Accuracy} & \multicolumn{1}{c}{RougeL} & \multicolumn{1}{c}{BERTscore} & \multicolumn{1}{c}{Accuracy} \\ \hline
EN & \textbf{.2260}                     & .8638                        & .5587                        & .1149                     & .8314                        & .6171                        & .2116                     & \textbf{.8650}                        & \textbf{.6762}                       \\
IT & \textbf{.1318}                     & .6934                        & .4054                        & .0741                     & .6513                        & .4619                        & .1227                     & \textbf{.6971}                        & \textbf{.5304}                       \\
ES & \textbf{.1740}                     & \textbf{.7231}                        & .4575                        & .1339                     & .6823                        & .4749                        & .1519                     & .7131                        & \textbf{.5142}                       \\
FR & \textbf{.1717}                     & \textbf{.6934}                        & .4942                        & .1299                     & .6807                        & .5506                        & .1129                     & .6819                        & \textbf{.6274}                       \\
DE & \textbf{.1512 }                    & \textbf{.6934 }                       & .5728                        & .0961                     & .6671                        & .5432                        & .1018                     & .6776                        & \textbf{.5975}                       \\ \hline
   & \multicolumn{3}{c|}{Llama 3.1 70B-Instruct}                                                         & \multicolumn{3}{c|}{Qwen2-72B-Instruct}                                                    & \multicolumn{3}{c}{Llama 3.1 405B-Instruct}                                                        \\ \cline{2-10} 
   & \multicolumn{1}{c}{RougeL} & \multicolumn{1}{c}{BERTscore} & \multicolumn{1}{c|}{Accuracy} & \multicolumn{1}{c}{RougeL} & \multicolumn{1}{c}{BERTscore} & \multicolumn{1}{c|}{Accuracy} & \multicolumn{1}{c}{RougeL} & \multicolumn{1}{c}{BERTscore} & \multicolumn{1}{c}{Accuracy} \\ \hline
EN & \textbf{.2437}                     & .8654                        & .7520                        & .1670                     & .8455                        & .7370                        & .2393                     & \textbf{.8669}                        & \textbf{.7532}                       \\
IT & .1416                     & .7000                        & .5688                        & .1145                     & .6747                        & \textbf{.5837}                        & \textbf{.1500}                     & \textbf{.7053}                        & .5716                       \\
ES & .1851                     & .7211                        & .5676                        & .1650                     & .7034                        & .5721                        & \textbf{.1858}                     & \textbf{.7265}                        & \textbf{.5766}                       \\
FR & .1554                     & \textbf{.7012}                        & \textbf{.6991}                        & .1430                     & .6934                        & .6940                        & \textbf{.1594}                     & .6952                        & .6799                       \\
DE & \textbf{.1263}                     & \textbf{.6928}                        & .6444                        & .1191                     & .6817                        & .6617                        & .1150                     & .6850                        & \textbf{.6642}                      
\end{tabular}\caption{Zero-shot results with machine translation.}\label{tab:resZeroMT}
\end{table*}

\subsection{Fine-tuning results}

This section reports results for the Llama3.1-8B fine-tuned model (Llama3.1-8B-FT).
We consider different training sizes (10K, 20K and the whole dataset (ALL)) and different data subsets with or without translated glosses during the training and testing, following the configurations shown in Table \ref{tab:ftconf}.
For 10K and 20K subsets, we maintain a balanced distribution between the generation and the multiple-choice tasks.
Specifically, we randomly select $x$ instances for each language for each task type, where $x$ corresponds to either 10K or 20K. Therefore, the whole training dataset consists of 100K and 200K for the 10K and 20K filtering, respectively. 

The FF and FT settings results for the models trained without machine translation are in Table \ref{tab:finetuning}. These results refer to the model fine-tuned only on the original glosses without machine translation.
It is essential to highlight that results for English change only when the training set varies since the machine translation does not affect the English test set. Generally, accuracy increases with the size of the training, except for German.
German has fewer training instances, and its performance is affected when the whole training set is used since our model is trained simultaneously in all languages.

\begin{table*}[!htb]
\small
\centering
\begin{tabular}{lrrrrrrrrr}
\multicolumn{1}{l|}{}   & \multicolumn{3}{c|}{\textbf{Llama3.1-8B-Instruct-FT / 10K}}                                                  & \multicolumn{3}{c|}{\textbf{Llama3.1-8B-Instruct-FT / 20K}}                                                  & \multicolumn{3}{c}{\textbf{Llama3.1-8B-Instruct-FT / ALL}}                                                  \\ \cline{2-10} 
\multicolumn{1}{l|}{}   & \multicolumn{1}{c}{RougeL} & \multicolumn{1}{c}{BERTscore} & \multicolumn{1}{c|}{Accuracy} & \multicolumn{1}{c}{RougeL} & \multicolumn{1}{c}{BERTscore} & \multicolumn{1}{c|}{Accuracy} & \multicolumn{1}{c}{RougeL} & \multicolumn{1}{c}{BERTscore} & \multicolumn{1}{c}{Accuracy} \\ \hline
\multicolumn{10}{c}{Configuration FF}                                                                                                                                                                                                                                                                         \\ \hline
\multicolumn{1}{l|}{EN} & .4584                     & .9021                        & \multicolumn{1}{r|}{.7788}   & .5346                     & .9139                        & \multicolumn{1}{r|}{.7889}   & .7392                     & .9466                        & .8067                       \\
\multicolumn{1}{l|}{IT} & .4739                     & .8068                        & \multicolumn{1}{r|}{.7580}   & .5649                     & .8350                        & \multicolumn{1}{r|}{.7881}   & .7452                     & .8920                        & .8234                       \\
\multicolumn{1}{l|}{ES} & .5166                     & .8362                        & \multicolumn{1}{r|}{.8156}   & .6098                     & .8680                        & \multicolumn{1}{r|}{.8329}   & .7649                     & .9191                        & .8694                       \\
\multicolumn{1}{l|}{FR} & .6108                     & .8629                        & \multicolumn{1}{r|}{.8937}   & .6831                     & .8904                        & \multicolumn{1}{r|}{.9253}   & .7923                     & .9258                        & .9163                       \\
\multicolumn{1}{l|}{DE} & .6484                     & .8659                        & \multicolumn{1}{r|}{.9043}   & .6904                     & .8802                        & \multicolumn{1}{r|}{.8870}   & .7106                     & .8890                        & .8739                       \\ \hline
\multicolumn{10}{c}{Configuration FT}                                                                                                                                                                                                                                                                         \\ \hline
\multicolumn{1}{l|}{EN} & .4584                     & .9021                        & \multicolumn{1}{r|}{.7788}   & .5346                     & .9139                        & \multicolumn{1}{r|}{.7889}   & .7392                     & .9466                        & .8067                       \\
\multicolumn{1}{l|}{IT} & .4139                     & .7856                        & \multicolumn{1}{r|}{.6648}   & .4920                     & .8100                        & \multicolumn{1}{r|}{.7032}   & .6436                     & .8581                        & .7499                       \\
\multicolumn{1}{l|}{ES} & .4046                     & .7942                        & \multicolumn{1}{r|}{.6416}   & .4600                     & .8123                        & \multicolumn{1}{r|}{.6551}   & .5689                     & .8482                        & .6924                       \\
\multicolumn{1}{l|}{FR} & .4484                     & .8026                        & \multicolumn{1}{r|}{.8156}   & .4913                     & .8189                        & \multicolumn{1}{r|}{.8323}   & .5581                     & .8405                        & .8464                       \\
\multicolumn{1}{l|}{DE} & .4476                     & .7884                        & \multicolumn{1}{r|}{.8346}   & .4705                     & .7969                        & \multicolumn{1}{r|}{.8741}   & .4837                     & .8021                        & .8395                      
\end{tabular}\caption{Evaluation results of the fine-tuned model trained on data without machine translation.}\label{tab:finetuning}
\end{table*}

Results of the model fine-tuned on machine-translated glosses are reported in Table \ref{tab:finetuningMT}.
The impact of machine translation during the training is minimal. If we consider tables \ref{tab:finetuning} and \ref{tab:finetuningMT}, where during the test, we do not use machine translation, we observe similar results. Some languages are more affected by the introduction of machine translation since these increase the number of instances, but not all languages are equally represented in the training data.

Introducing translated instances in the test set impacts performance for many reasons. First, the number of instances to test increases, and we may introduce instances with more polysemy, especially in the multiple-choice benchmark, in which we add new glosses and potentially introduce more alternatives.

\begin{table*}[!htb]
\small
\centering
\begin{tabular}{lrrrrrrrrr}
\multicolumn{1}{l|}{}   & \multicolumn{3}{c|}{\textbf{Llama3.1-8B-Instruct-FT / 10K}}                                                  & \multicolumn{3}{c|}{\textbf{Llama3.1-8B-Instruct-FT / 20K}}                                         & \multicolumn{3}{c}{\textbf{Llama3.1-8B-Instruct-FT / ALL}}                                                  \\ \cline{2-10} 
\multicolumn{1}{l|}{}   & \multicolumn{1}{c}{RougeL} & \multicolumn{1}{c}{BERTscore} & \multicolumn{1}{c|}{Accuracy} & \multicolumn{1}{c}{RougeL} & \multicolumn{1}{c}{BERTscore} & \multicolumn{1}{c|}{Accuracy} & \multicolumn{1}{c}{RougeL} & \multicolumn{1}{c}{BERTscore} & \multicolumn{1}{c}{Accuracy} \\ \hline
\multicolumn{10}{c}{Configuration TF}                                                                                                                                                                                                                                                                         \\ \hline
\multicolumn{1}{l|}{EN} & .4586                     & .9018                        & \multicolumn{1}{r|}{.7776}   & .5493                     & .9163                        & \multicolumn{1}{r|}{.7877}   & .7446                     & .9477                        & .8224                       \\
\multicolumn{1}{l|}{IT} & .4154                     & .7880                        & \multicolumn{1}{r|}{.7410}   & .5336                     & .8245                        & \multicolumn{1}{r|}{.7737}   & .7128                     & .8815                        & .8234                       \\
\multicolumn{1}{l|}{ES} & .4175                     & .8051                        & \multicolumn{1}{r|}{.8204}   & .4981                     & .8308                        & \multicolumn{1}{r|}{.8367}   & .6972                     & .8982                        & .8655                       \\
\multicolumn{1}{l|}{FR} & .5666                     & .8481                        & \multicolumn{1}{r|}{.9072}   & .6530                     & .8745                        & \multicolumn{1}{r|}{.8914}   & .7779                     & .9205                        & .9186                       \\
\multicolumn{1}{l|}{DE} & .5678                     & .8373                        & \multicolumn{1}{r|}{.9043}   & .6669                     & .8721                        & \multicolumn{1}{r|}{.8478}   & .7079                     & .8860                        & .8652                       \\ \hline
\multicolumn{10}{c}{Configuration TT}                                                                                                                                                                                                                                                                         \\ \hline
\multicolumn{1}{l|}{EN} & .4586                     & .9018                        & \multicolumn{1}{r|}{.7776}   & .5493                     & .9163                        & \multicolumn{1}{r|}{.7877}   & .7446                     & .9477                        & .8224                       \\
\multicolumn{1}{l|}{IT} & .3825                     & .7767                        & \multicolumn{1}{r|}{.6621}   & .4920                     & .8096                        & \multicolumn{1}{r|}{.6967}   & .6703                     & .8662                        & .7575                       \\
\multicolumn{1}{l|}{ES} & .3790                     & .7877                        & \multicolumn{1}{r|}{.6499}   & .4469                     & .8085                        & \multicolumn{1}{r|}{.6744}   & .6460                     & .8729                        & .7079                       \\
\multicolumn{1}{l|}{FR} & .4623                     & .8100                        & \multicolumn{1}{r|}{.8233}   & .5290                     & .8301                        & \multicolumn{1}{r|}{.8105}   & .6497                     & .8716                        & .8399                       \\
\multicolumn{1}{l|}{DE} & .4573                     & .7897                        & \multicolumn{1}{r|}{.8247}   & .5143                     & .8090                        & \multicolumn{1}{r|}{.8222}   & .5553                     & .8250                        & .7852                      
\end{tabular}\caption{Evaluation results of the fine-tuned model trained on data with machine translation.}\label{tab:finetuningMT}
\end{table*}

To compare our results with the ones proposed in \cite{pasini-etal-xl-wsd-2021}, we must consider the test set with translated glosses to cover all the instances in the original dataset.
We only consider configurations on the multiple-choice benchmark since the XL-WSD evaluation metric (F1) is based on sense prediction. The quality of the definition generation cannot be compared with other systems since the other WSD approaches do not generate a sense definition.
Since the multiple-choice task is also performed as a generative problem, we need to transform the answer provided by the LLM into the BabelNet sense id used by the XL-WSD scoring tool.
This process is not trivial and requires several steps:
\begin{enumerate}
    \item We extract the choice from the LLM answer using a regular expression;
    \item The choice is used to extract the gloss from the instruction provided to the LLM;
    \item The gloss is used to retrieve the sense id by searching over the glosses of the possible senses of the target word;
    \item The multiple-choice dataset does not contain instances of monosemic words. We select the only available sense id as the prediction in these cases.
\end{enumerate}

\begin{table*}[!htb]
\centering
\begin{tabular}{l|rrrrr|r}
                           & \multicolumn{1}{c}{\textbf{EN}} & \multicolumn{1}{c}{\textbf{IT}} & \multicolumn{1}{c}{\textbf{ES}} & \multicolumn{1}{c}{\textbf{FR}} & \multicolumn{1}{c|}{\textbf{DE}} & \multicolumn{1}{c}{\textbf{AVG}} \\ \hline
XLMR-Large                 & .7628                          & .7766                          & .7585                          & .8388                          & .8318                           & .7937                           \\
XLMR-Base                  & .7450                          & .7673                          & .7655                          & .8233                          & .8213                           & .7845 \\
BERT-L	& .7677 & - & - & - & - & - \\
BERT-M	& -	& .7616 & .7466 &	.8164 &	.8063 & - \\
LS-BERT &	-	& .7388	& .7477	& .8078	& .8213 & - \\
QInterf & .8010$\diamond$ & .7980 & \textbf{.7900} & .8500 & .8500 & -
\\ \hline
\textit{Llama3.1-8B-FT / ALL} & \textbf{.8652}                 & \textbf{.8205}                 & .7769                 & \textbf{.8836}                 & \textbf{.8898}                  & \textbf{.8472}                  \\ \hline
Gemma2-9B (0-shot)         & .7343                          & .6295                          & .5997                          & .7370                          & .8016                           & .7004                           \\
Llama3.1-70B (0-shot)      & .7988                          & .6646                          & .6424                          & .7888                          & .8237                           & .7437                           \\
Llama3.1-405B (0-shot)     & .8040                          & .6699                          & .6509                          & .7759                          & .8329                           & .7467                          
\end{tabular}\caption{XL-WSD results. $\diamond$ The QInterf system is evaluated on a different portion of the English dataset that does not include data from SemEval-2010.}\label{tab:resXLWSD}
\end{table*}

Analyzing Tables \ref{tab:finetuning} and \ref{tab:finetuningMT}, we observe that when the translated test set is used, English, Italian and Spanish achieve the best accuracy with the whole machine-translated training set. Conversely, French and German perform best with the training set without machine translation: French performs best with the whole training set, while German with 20K instances. We decide to consider the \texttt{TT} setting for all the languages since, in a real scenario, using different training for each language is not feasible.

In Table \ref{tab:resXLWSD}, we report the F1 computed using the official scoring tool released by the XL-WSD creator. The table also shows the results of current best systems: \textit{XLMR-Large} and \textit{XLMR-Base} based on \cite{conneau-etal-2020-unsupervised}, supervised approaches based on BERT. Moreover, we report several BERT-based systems: BERT-L is the large model specific for English, BERT-M is the multilingual model and LS-BERT is the language-specific model. Finally, we add a recent system QInterf \cite{zhang2024quantum} based on a quantum interference model that calculates the probability that the target word belongs to a superposition state representing the multiple glosses of the same word sense.
Table \ref{tab:resXLWSD} also shows the best LLMs in the zero-shot setting and our fine-tuned model (\textit{Llama-8B-FT / ALL}).
Our fine-tuned model performs best for all languages, with a remarkable result of \textbf{.8652} for English. The results allow some interesting considerations:
\begin{itemize}
    \item LLMs in zero-shot learning are not able to overcome the baseline models except large models that provide better results for English and German;
    \item all LLMs in zero-shot setting show poor performance for Italian and Spanish;
    \item a medium model (Llama3.1-8B) fine-tuned on training data provides impressive results and always overcomes large models and baselines, resulting in state-of-the-art performance.
\end{itemize}

\section{Conclusions and Future Work}\label{sec:conclusions}

In this work, we investigate the performance of several LLMs in solving the WSD task. For that purpose, we extend an existing benchmark (XL-WSD) to support two new subtasks: 1) given a word occurrence in a sentence, the LLM must provide the correct definition; 2) given a word occurrence in a sentence and a set of predefined meanings, the LLM must select the correct on. To build our benchmark, we exploit the XL-WSD dataset and BabelNet. Moreover, we use training data available in XL-WSD for fine-tuning and LLM based on Llama3.1-8B.
Results show that LLMs can provide good performance in zero-shot learning but are not able to overcome current state-of-the-art approaches. The best performances are obtained by large models, while medium ones provide poor results.
However, the fine-tuned model with a medium number of parameters is able to overcome all the models, including the current state-of-the-art approaches. The fine-tuned model can achieve an impressive accuracy of \textbf{.8472} averaging all languages, and a remarkable accuracy of \textbf{.8652} in English.

\section{Limitations}

The current version of our work presents some limitations summarized in the following points:
\begin{enumerate}
    \item Not all the languages presented in XL-WSD are taken into account. We focused on languages with adequate resources to ensure a robust evaluation pipeline. However, we plan to extend the analysis to underrepresented languages.
    \item The few-shot approach is not considered. We have decided to exclude this approach to reduce the complexity of the paper. For the same reason, we did not consider a multi-prompt evaluation. Nonetheless, we are aware of their potential and will explore them in the future. 
    \item The fine-tuning approach involves only one model with a medium number of parameters, excluding larger models used in the zero-shot evaluation. This choice was made to investigate strategies that can be implemented on affordable hardware.
    \item While the exclusion of ChatGPT may be seen as a limitation, we aim to promote the use of open models to improve reproducibility and transparency in research. In line with this, we also use Llama3.1-405b, which provides performance similar to ChatGPT-4o in several state-of-the-art benchmarks.
\end{enumerate}

\section{Ethical considerations}

Our work is heavily based on pre-trained LLMs developed by external organizations. The pre-training procedure was performed without our supervision, and the datasets used for pre-training and fine-tuning were also not checked. Therefore, the models may produce inaccurate or biased results that reflect the potential issues present in the original training data.

To reduce inaccuracies, human experts manually checked the prompt templates for each language. These experts participated voluntarily and were fully informed about our research objectives and the use of the data they checked, ensuring transparency in their involvement and contributions.

\section{Resources}

We release several resources:
\begin{itemize}
    \item Training and testing data of our benchmark are available on Zenodo\footnote{\url{https://zenodo.org/records/15007563}}
    \item The outputs of LLMs are also available on Zenodo
    \item The source code is available on GitHub\footnote{\url{https://github.com/swapUniba/LLM-wsd}}
    \item Fine-tuned models are available on HuggingFace\footnote{\url{https://huggingface.co/collections/swap-uniba/llm-wsd-67c9a90da3b4918bf7ea1037}}
\end{itemize}

\section*{Acknowledgments}
We acknowledge the support of the PNRR project FAIR - Future AI Research (PE00000013), Spoke 6 - Symbiotic AI (CUP H97G22000210007) under the NRRP MUR program funded by the NextGenerationEU.

\bibliographystyle{unsrtnat}
\bibliography{references}

\appendix

\section{Appendix A}\label{sec:appendixA}

This appendix reports some detailed statistics about the dataset.
Table \ref{taba:notransl} shows the number of instances for each kind of benchmark and language without glosses translation. Meanwhile, Table \ref{taba:transl} reports the same statistics when machine translation is exploited to create missing glosses.

\begin{table}[!h]
\centering
\begin{tabular}{lllll}
\multicolumn{5}{c}{\textbf{Without machine translation}}                                                                                                             \\ \hline
\multicolumn{1}{l|}{}   & \multicolumn{2}{c|}{generation}                                  & \multicolumn{2}{c}{selection}                                   \\ \cline{2-5} 
\multicolumn{1}{l|}{}   & \multicolumn{1}{c}{training set} & \multicolumn{1}{c|}{test set} & \multicolumn{1}{c}{training set} & \multicolumn{1}{c}{test set} \\ \hline
\multicolumn{1}{l|}{EN} & 565,831                          & \multicolumn{1}{l|}{6,757}    & 421,213                          & 6,605                        \\
\multicolumn{1}{l|}{IT} & 242,343                          & \multicolumn{1}{l|}{1,673}    & 192,962                          & 1,529                        \\
\multicolumn{1}{l|}{ES} & 216,317                          & \multicolumn{1}{l|}{1,248}    & 153,817                           & 1,041                          \\
\multicolumn{1}{l|}{FR} & 121,014                          & \multicolumn{1}{l|}{539}      & 63,429                           & 442                          \\
\multicolumn{1}{l|}{DE} & 58,925                           & \multicolumn{1}{l|}{263}      & 30,370                            & 230                         
\end{tabular}\caption{Number of instances for each language without machine translation.}\label{taba:notransl}
\end{table}

\begin{table}[!h]
\centering
\begin{tabular}{lllll}
\multicolumn{5}{c}{\textbf{With machine translation}}                                                                                                                      \\ \hline
\multicolumn{1}{l|}{}   & \multicolumn{2}{c|}{generation}                                        & \multicolumn{2}{c}{selection}                                   \\ \cline{2-5} 
\multicolumn{1}{l|}{}   & \multicolumn{1}{c}{training set} & \multicolumn{1}{c|}{test set}       & \multicolumn{1}{c}{training set} & \multicolumn{1}{c}{test set} \\ \hline
\multicolumn{1}{l|}{EN} & \textit{565,831}                 & \multicolumn{1}{l|}{\textit{6,757}} & \textit{421,213}                 & \textit{6,605}               \\
\multicolumn{1}{l|}{IT} & 308,903                          & \multicolumn{1}{l|}{1,888}          & 271,864                          & 1,823                        \\
\multicolumn{1}{l|}{ES} & 345,258                          & \multicolumn{1}{l|}{1,601}          & 319,156                          & 1,554                        \\
\multicolumn{1}{l|}{FR} & 157,787                          & \multicolumn{1}{l|}{812}            & 113,307                           & 781                          \\
\multicolumn{1}{l|}{DE} & 73,871                           & \multicolumn{1}{l|}{415}            & 45,381                           & 405                         
\end{tabular}\caption{Number of instances for each language with machine translation.}\label{taba:transl}
\end{table}

\section{Appendix B}\label{sec:appendixB}

This appendix reports prompts in languages other than English. Native-language speakers or translation experts have checked all prompts.
Some sentences have grammatical issues since the XL-WSD dataset could contain data obtained from machine-translated corpora.
Prompts for Italian are in tables \ref{tab:promptGenIt} and \ref{tab:promptSelIt}.
Prompts for Spanish are in tables \ref{tab:promptGenEs} and \ref{tab:promptSelEs}.
Prompts for French are in tables \ref{tab:promptGenFr} and \ref{tab:promptSelFr}.
Prompts for German are in tables \ref{tab:promptGenDe} and \ref{tab:promptSelDe}.
 
\begin{table}[!h]
\centering
\begin{tabular}{|p{7cm}|}
\hline
\multicolumn{1}{|c|}{\textbf{Prompt template (generation)}} \\
\hline
Fornisci una breve definizione della parola "$w_i$" nella frase data in input. Genera solo la definizione. Input: "$S_k$"  \\
\hline
\multicolumn{1}{|c|}{\textbf{Italian prompt}}  \\
\hline
Fornisci una breve definizione della parola "sforzo" nella frase data in input. Genera solo la definizione. Input: "Che sforzo fate per valutare i risultati del vostro programme?" \\
\hline
\end{tabular}\caption{Prompt for the Italian generation benchmark.}\label{tab:promptGenIt}
\end{table}

\begin{table}[!h]
\centering
\begin{tabular}{|p{7cm}|}
\hline
\multicolumn{1}{|c|}{\textbf{Prompt template (multiple choice)}} \\
\hline
Data la parola "$w_i$" nella frase in input, scegli il significato corretto tra i seguenti:
$C_i$.
Genera solo il numero dell'opzione selezionata. Input: "$S_k$"  \\
\hline
\multicolumn{1}{|c|}{\textbf{Italian prompt}}  \\
\hline
Data la parola "valutare" nella frase in input, scegli il significato corretto tra i seguenti:\newline
1) Esaminare o ascoltare (prove o un intero caso) per via giudiziaria.\newline
2) Fare la stima commerciale di qlco.\newline
3) Assegnare un valore a.\newline
4) Ritenere dopo valutazione.\newline
5) Apprezzare, tenere in grande stima.\newline
6) Avere una certa opinione di qualcuno.\newline
Genera solo il numero dell'opzione selezionata. \newline
Input: "Che sforzo fate per valutare i risultati del vostro programme?"
\\\hline
\end{tabular}\caption{Prompt for the Italian multiple choice benchmark.}\label{tab:promptSelIt}
\end{table}

\begin{table}[!h]
\centering
\begin{tabular}{|p{7cm}|}
\hline
\multicolumn{1}{|c|}{\textbf{Prompt template (generation)}} \\
\hline
Proporciona una definición breve de la palabra "$w_i$" en la frase dada en entrada. Genera solo la definición. Input: "$S_k$"  \\
\hline
\multicolumn{1}{|c|}{\textbf{Spanish prompt}}  \\
\hline
Proporciona una definición breve de la palabra "reducido" en la frase dada en entrada. Genera solo la definición. Input: "¿ Mida su relación con el absentismo reducido, el volumen de negocios, los accidentes y las quejas, y con la mejora de la calidad y la producción?" \\
\hline
\end{tabular}\caption{Prompt for the Spanish generation benchmark.}\label{tab:promptGenEs}
\end{table}

\begin{table}[!h]
\centering
\begin{tabular}{|p{7cm}|}
\hline
\multicolumn{1}{|c|}{\textbf{Prompt template (multiple choice)}} \\
\hline
Dada la palabra "$w_i$" en la frase de entrada, elija el significado correcto entre los siguientes:
$C_i$.
Genera solo el número de la opción seleccionada. Input: "$S_k$" \\
\hline
\multicolumn{1}{|c|}{\textbf{Spanish prompt}}  \\
\hline
Dada la palabra "esfuerzo" en la frase de entrada, elija el significado correcto entre los siguientes:\\
1) Actividad seria y consiente para hacer o lograr algo.\\
2) utilización de la fuerza y de otros medios por encima de lo normal con el fin de lograr un determinado objetivo\\
3) Ejercicio intenso o violento.\\
4) Enérgico intento de conseguir algo.\\
5) Intento que requiere un esfuerzo para conseguir un objetivo.\\

Genera solo el número de la opción seleccionada.\\

Input: "¿ Qué esfuerzo hace para evaluar los resultados de su programa?"\\
\\\hline
\end{tabular}\caption{Prompt for the Spanish multiple choice benchmark.}\label{tab:promptSelEs}
\end{table}

\begin{table}[!h]
\centering
\begin{tabular}{|p{7cm}|}
\hline
\multicolumn{1}{|c|}{\textbf{Prompt template (generation)}} \\
\hline
Donnez une brève définition du mot "$w_i$" dans la phrase d'entrée donnée. Ne donnez que la définition. Input: "$S_k$"  \\
\hline
\multicolumn{1}{|c|}{\textbf{French prompt}}  \\
\hline
Donnez une brève définition du mot "production" dans la phrase d'entrée donnée. Ne donnez que la définition. Input: "Mesurez -vous son rapport à la réduction de l'absentéisme, de chiffre d'affaires, des accidents et des griefs, ainsi qu'à l'amélioration de la qualité et de la production?" \\
\hline
\end{tabular}\caption{Prompt for the French generation benchmark.}\label{tab:promptGenFr}
\end{table}

\begin{table}[!h]
\centering
\begin{tabular}{|p{7cm}|}
\hline
\multicolumn{1}{|c|}{\textbf{Prompt template (multiple choice)}} \\
\hline
Étant donné le mot "$w_i$" dans la phrase saisie, choisissez la signification correcte parmi les suivantes: $C_i$.
Ne donnez que le numéro de l'option sélectionnée. Input: "$S_k$" \\
\hline
\multicolumn{1}{|c|}{\textbf{French prompt}}  \\
\hline
Étant donné le mot "essayez" dans la phrase saisie, choisissez la signification correcte parmi les suivantes:\\
1) Mettre à l’essai.\\
2) S'exercer à faire ou à effectuer quelque chose.\\
3) Tester l'apparence et la taille de (un vêtement) en le portant.\\

Ne donnez que le numéro de l'option sélectionnée.\\

Input: "Lorsque de améliorations sont recommandées dans les conditions de travail - comme l'éclairage, les salles de repos, les restaurants, la climatisation - essayez -vous de déterminer leur efficacité sur la productivité?"\\
\\\hline
\end{tabular}\caption{Prompt for the French multiple choice benchmark.}\label{tab:promptSelFr}
\end{table}

\begin{table}[!h]
\centering
\begin{tabular}{|p{7cm}|}
\hline
\multicolumn{1}{|c|}{\textbf{Prompt template (generation)}} \\
\hline
Geben Sie eine kurze Definition des Wortes "$w_i$" in dem gegebenen Satz an. Erzeugen Sie nur die Definition. Input: "$S_k$"  \\
\hline
\multicolumn{1}{|c|}{\textbf{German prompt}}  \\
\hline
Geben Sie eine kurze Definition des Wortes "Ziele" in dem gegebenen Satz an. Erzeugen Sie nur die Definition. Input: "Erreicht sie diese Ziele?" \\
\hline
\end{tabular}\caption{Prompt for the German generation benchmark.}\label{tab:promptGenDe}
\end{table}

\begin{table}[!h]
\centering
\begin{tabular}{|p{7cm}|}
\hline
\multicolumn{1}{|c|}{\textbf{Prompt template (multiple choice)}} \\
\hline
Wählen Sie für das Wort "$w_i$" im Eingabesatz die richtige Bedeutung aus den folgenden Angaben: $C_i$.
Erzeugt nur die Nummer der ausgewählten Option. Input: "$S_k$" \\
\hline
\multicolumn{1}{|c|}{\textbf{German prompt}}  \\
\hline
Wählen Sie für das Wort "Wahl" im Eingabesatz die richtige Bedeutung aus den folgenden Angaben:\\
1) Die Auswahl von etwas aus mehreren Möglichkeiten oder Alternativen.\\
2) Ein Stimmzettel, auch Wahlzettel, ist ursprünglich ein Zettel, auf dem der Wähler seine Wahl handschriftlich kundtun kann.\\
3) Weiler in Russland\\
4) Eine Wahl im Sinne der Politikwissenschaft ist ein Verfahren in Staaten, Gebietskörperschaften und Organisationen zur Bestellung einer repräsentativen Person oder mehrerer Personen als entscheidungs- oder herrschaftsausübendes Organ.\\

Erzeugt nur die Nummer der ausgewählten Option.\\

Input: "Stellen Sie bei Verhandlungen mit Ihrer Gewerkschaft sicher, dass die Mitarbeiter die Wahl zwischen neuen Leistungen und ihren Cents pro Stunde Lohnkosten haben."
\\\hline
\end{tabular}\caption{Prompt for the German multiple choice benchmark.}\label{tab:promptSelDe}
\end{table}

\end{document}